\documentclass[10pt,twocolumn,letterpaper]{article}

\usepackage{iccv}
\usepackage{times}
\usepackage{epsfig}
\usepackage{graphicx}
\usepackage{amsmath}
\usepackage{amssymb}

\usepackage{multirow}
\usepackage{makecell}
\usepackage{chngpage}

\usepackage[breaklinks=true,bookmarks=false]{hyperref}

\iccvfinalcopy 

\def\iccvPaperID{9342} 

\ificcvfinal\fi

\begin{document}

\title{Prior to Segment: Foreground Cues for Weakly Annotated Classes \\ in Partially Supervised Instance Segmentation}

\author{David Biertimpel$^{1,2}$\thanks{This paper is the product of work during an internship at TomTom.}~~~~Sindi Shkodrani$^{2}$~~~~Anil S. Baslamisli$^{1}$~~~~Nóra Baka$^{2}$\\[1mm]
\normalsize $^{1}$University of Amsterdam~~~~~$^{2}$TomTom\\[1mm]
{\tt\small david.biertimpel@protonmail.com~~~~a.s.baslamisli@uva.nl}\\[0.0mm]
{\tt\small \{sindi.shkodrani, nora.baka\}@tomtom.com}
}

\maketitle
\ificcvfinal\thispagestyle{empty}\fi

\begin{abstract}
Instance segmentation methods require large datasets with expensive and thus limited instance-level mask labels. Partially supervised instance segmentation aims to improve mask prediction with limited mask labels by utilizing the more abundant weak box labels. In this work, we show that a class agnostic mask head, commonly used in partially supervised instance segmentation, has difficulties learning a general concept of foreground for the weakly annotated classes using box supervision only. To resolve this problem we introduce an object mask prior (OMP) that provides the mask head with the general concept of foreground implicitly learned by the box classification head under the supervision of all classes. This helps the class agnostic mask head to focus on the primary object in a region of interest (RoI) and improves generalization to the weakly annotated classes. We test our approach on the COCO dataset using different splits of strongly and weakly supervised classes. Our approach significantly improves over the Mask R-CNN baseline and obtains competitive performance with the state-of-the-art, while offering a much simpler architecture. \footnote {Code is available at: \url{https://github.com/dbtmpl/OPMask}}
\end{abstract}


\section{Introduction}

\begin{figure}
\begin{center}
\includegraphics[width=1.0\columnwidth]{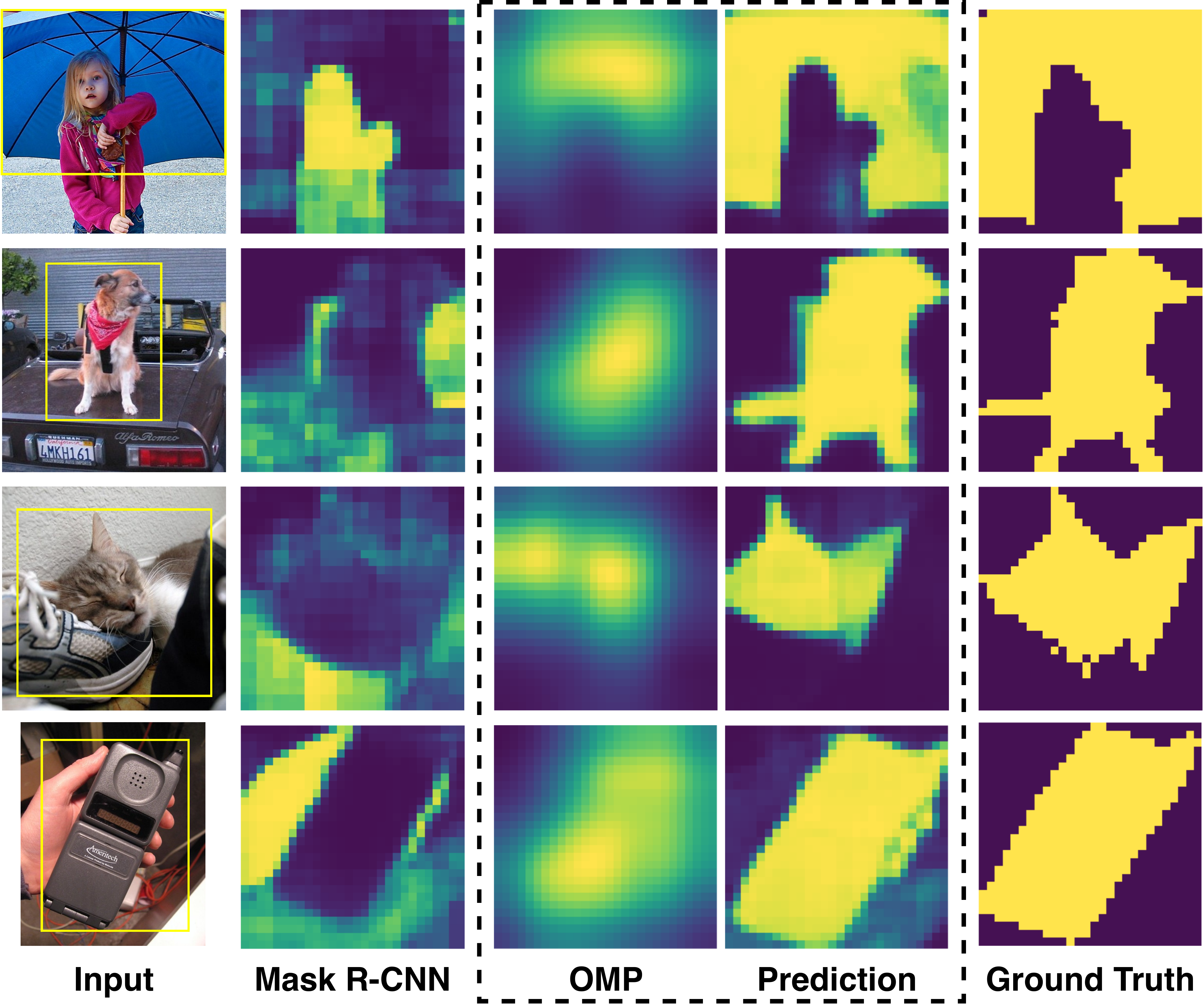}
\end{center}
   \caption{Our object mask prior (OMP) provides foreground cues to the mask head highlighting the primary instance in ambiguous RoIs. OPMask is able to resolve ambiguous constellations and segment the correct instance, while our Mask R-CNN baseline fails to do so. Yellow pixels indicate the foreground (primary instance) in the RoI, blue pixels the background.}
\label{fig:fg_bg}
\end{figure}

 Instance segmentation is an essential task in computer vision with applications ranging from autonomous vehicles to robotics and medical imaging \cite{blendmask, boundary_pres, mask_rcnn, centermask, panet, deep_snake}. A major contributor to the success of recent instance segmentation methods is the availability of large-scale datasets with numerous instance-level mask labels \cite{cityscapes, pascal_voc, lvis, coco, isaid}. A major problem with mask labels is that their acquisition is rather time-consuming at $\sim 67$ seconds per instance \cite{exp_mask_anno}. Conversely, weak labels (\wrt instance masks) such as bounding boxes ($\sim 10.5$ seconds) and image-level labels ($\sim 1$ second) allow for much more efficient annotation \cite{exp_mask_anno}. Nonetheless, for classes without sufficient mask labels, conventional instance segmentation methods perform poorly and tend to generate mask predictions that are perforated, do not cover the entire object or are completely missing it \cite{ltse, shapemask}. To improve mask predictions for classes with no mask labels available, recent research has focused on addressing the problem in a partially supervised learning setting \cite{cpmask, ltse, shapemask, shapeprop}, where all classes are annotated with box labels while only a subset of these classes also carry instance mask labels. The goal in partially supervised instance segmentation is to use the abundant but weak box labels in conjunction with the strong but limited mask labels to predict better instance masks for the weakly (box) annotated classes.

In current methods, the task of generalizing mask predictions for weak classes is either achieved with meta-learning of class aware weights \cite{ltse} or with a class agnostic mask head \cite{cpmask, shapemask, shapeprop}. In the latter case, instead of predicting a mask per class, each pixel in the RoI is classified into either foreground or background. Thus, the class agnostic mask head faces the challenge of having to learn a general concept of foreground in order to generalize to unseen object classes. Nonetheless, this often fails, even if abundant box labels are provided for the weak classes. 

In this paper, we identify that the problem originates, on the one hand, from the ambiguous constellations between object instances, where pixels of one instance appear in the bounding box of the other. Thus, the actual foreground becomes ambiguous to the mask head when the RoI contains multiple and possibly overlapping instances. See Figure \ref{fig:fg_bg} for examples. On the other hand, instances of weak classes that appear in the background of a RoI during training are actively learned as background. This hurts generalization to weak classes that frequently interact with other supervised classes. To address these problems, we introduce an object mask prior (OMP) that highlights the correct foreground in each RoI. This helps the mask head to resolve ambiguous constellations, learn a more general concept of the foreground, and generalize it to weak classes.

Recent works have demonstrated that shape priors are beneficial inductive biases that steer models towards more stable mask predictions. For example, ShapeMask \cite{shapemask} creates a knowledge base of shape priors by applying k-means to the ground-truth masks, whereas ShapeProp \cite{shapeprop} creates priors by using pixel-wise multiple instance learning on bounding boxes. Although these priors help to generalize to weak classes, they do not explicitly address the problems mentioned above.

Conversely, our prior is explicitly optimized to highlight the foreground in a RoI using the box supervision from all classes. This is achieved by exploiting the fact that the box classification head naturally learns to identify the primary class in a RoI. As the box head receives labels for all classes in the partially supervised setting, the box features capture a general concept of foreground. To reveal this foreground, we use class activation maps (CAMs) \cite{cams}, which are coarse localization maps indicating the most discriminative image regions detected by the model. Therefore, given a correct classification, CAMs are expected to highlight foreground areas corresponding to the primary RoI class. 

Unlike other methods that introduce separate modules for prior creation, we natively embed the OMP into our model in an end-to-end manner, without introducing any additional architectural overhead. Besides using box supervision from all classes, our prior is able utilize mask gradients originated from the limited mask labels to increase its spatial extent. To realize our prior, we embed it in the Mask R-CNN meta architecture and name our overall model OPMask (Object Prior Mask R-CNN). Our main contributions are the following:
\begin{itemize}
    \item We identify two fundamental problems in partially supervised instance segmentation: First, instances of weak classes appearing in the background of a mask supervised RoI during training are learned as background by the model. Second, in ambiguous RoIs containing multiple and possibly overlapping instances, the mask head has difficulties finding the foreground. 
    \item We introduce an object mask prior (OMP) in the mask head to solve the above identified problems. The prior highlights the foreground across all classes by leveraging the information from the box head.
    \item On the COCO dataset \cite{coco}, OPMask significantly improves over our Mask R-CNN baseline by $13.0$ AP. Compared with the prior state-of-the-art, we improve over ShapeMask \cite{shapemask} and ShapeProp \cite{shapeprop} and achieve competitive results against CPMask \cite{cpmask} while using a simpler architecture.
\end{itemize}

\section{Related Work}
\textbf{Instance segmentation} aims to segment every object instance in a scene. Detection based approaches \cite{boundary_pres, mask_rcnn, centermask, panet}, which add a mask prediction network to existing detection models, represent the current state-of-the-art. Mask R-CNN \cite{mask_rcnn} extends the two stage detection network Faster R-CNN \cite{Faster_R_CNN} being the first to introduce a multi-task loss combining detection and mask gradients. Mask R-CNN is a strong baseline and often used as a meta-architecture due to its extensibility. Contour based approaches \cite{marcos2018learning, deep_snake, xu2019explicit} segment objects by refining a sequence of vertices to match the object shape. Bottom-up approaches group pixels to generate instance masks \cite{arnab2017pixelwise, liu2017sgn, cluster_group}. As these approaches need large datasets with pixel-wise supervision, they are not suited for the partially supervised task.

\begin{figure*}
\begin{center}
\includegraphics[width=0.9\linewidth]{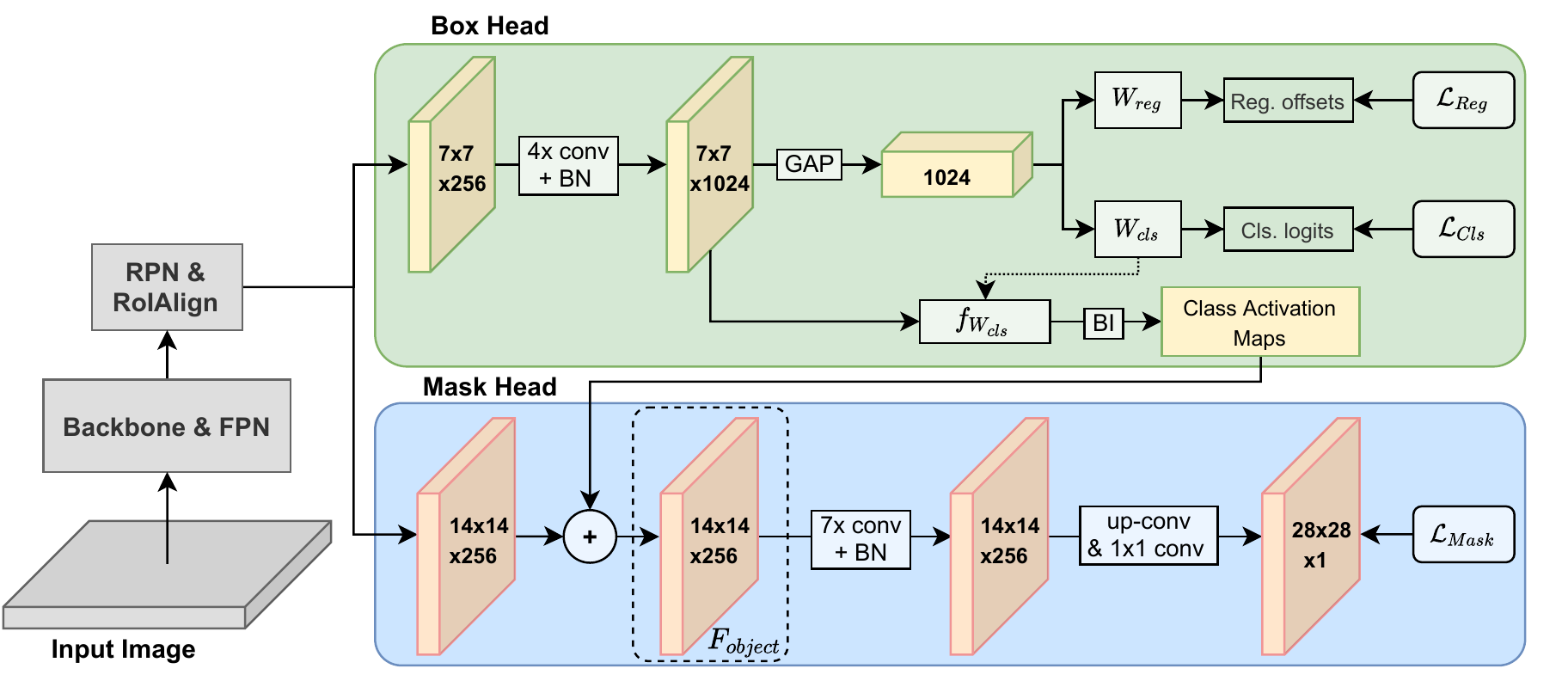}
\end{center}
   \caption{Overall architecture. The box head generates our OMP which is added to features entering the mask head to create object aware features $F_{\textit{object}}$. The mask head then uses $F_{\textit{object}}$ to predict instance masks.}
\label{fig:architecture}
\end{figure*}

\textbf{Partially supervised instance segmentation.} In partially supervised instance segmentation, a subset of classes is strongly annotated with box and mask supervision, while the remaining classes carry only weak box labels. The goal is to use the box labels in conjunction with the limited masks to predict instance masks for all classes. 

The pioneering approach by Hu \etal \cite{ltse} augments a Mask R-CNN with a weight transfer function that learns a mapping from box to mask weights, introducing a class aware mask head capturing a representation for all classes. 

Kuo \etal introduce ShapeMask \cite{shapemask} that creates a knowledge base of shape priors by applying k-means to the available ground-truth masks. A box embedding gives rise to a linear combination of the k-means centroids generating a shape prior that is further refined into an instance mask. ShapeMask bases its prior solely on the limited mask labels. In contrast, we use box labels of all available classes and use mask labels for refinement. 

ShapeProp \cite{shapeprop} uses pixel-wise multiple instance learning (MIL) on bounding boxes to create a saliency heatmap, which is further processed leading to a more expressive shape activation map. Both ShapeProp and OPMask utilize box labels to generate a prior for mask prediction. ShapeProp introduces two separate modules to generate and refine their prior. On the other hand, we take advantage of the fact that the box head implicitly learns a concept of foreground. Thus, we design our model to leverage the features that are already made available by the box head. This way we do not introduce any architectural overhead. 

Finally, Fan \etal \cite{cpmask} learn the underlying shape and appearance commonalities between instance masks that should generalize to weak classes. The shape commonalities are learned by a boundary prediction head, while the appearance commonalities are enforced by an attention based affinity parsing module. Besides learning commonalities that aid generalization, we also identify that a major problem lies in ambiguous RoIs and the mask head having difficulties to learn a general concept of foreground. To address this, we utilize our OMP, which highlights the foreground of a RoI to resolve ambiguous constellations and help generalize to weak classes.

\textbf{Weakly supervised instance segmentation} approaches solely rely on weak labels such as bounding boxes or images level labels \cite{wi_inter_pix_ahn, wi_consistent_arun, wi_seq_label, wi_bb_tight, wi_sm_sdi_refine_box, wi_where_masks_img_level, wi_class_peak, wi_inst_act}. Models using image-level labels \cite{wi_inter_pix_ahn, wi_seq_label,  wi_where_masks_img_level, wi_class_peak, wi_inst_act} mostly use CAM based image-wide localization information to assist instance mask prediction. Zhou \etal \cite{wi_class_peak} use the peaks of a class response map to detect and segment instances. Ge \etal \cite{wi_seq_label} refine object attention maps using multi-task network heads sharing the same backbone. Both Laradji \etal \cite{wi_where_masks_img_level} and Ahn \etal \cite{wi_inter_pix_ahn} create and refine pseudo masks which are later used to train a Mask R-CNN \cite{mask_rcnn}. Setups where only image-level labels are available require the introduction of complex refinement modules. Conversely, in our setting, we rely on mask gradients that are already available in the model to improve our OMP. 

Less work has been done using box supervision \cite{wi_bb_tight, wi_sm_sdi_refine_box}. Hsu \etal \cite{wi_bb_tight} employ a Mask R-CNN like architecture, where the mask head uses a MIL objective. Khoreva \etal \cite{wi_sm_sdi_refine_box} use GrabCut \cite{grabcut} to create pseudo ground truths to train a separate segmentation model. Instead of using box pixels to predict masks, we use CAMs to extract the foreground information in the box features to create our OMP.


\section{Method}
In partially supervised instance segmentation, a conventional Mask R-CNN with a class agnostic mask head fails to predict reliable instance masks for certain weak classes, as demonstrated in Figures \ref{fig:fg_bg} and \ref{fig:prior_and_preds}, and as discussed in the introduction. 
To address this, we propose OPMask which introduces an object mask prior (OMP) that captures foreground cues for all classes in the dataset (i.e. generalized foreground). OPMask follows the design of a Mask R-CNN \cite{mask_rcnn} with a ResNet \cite{resnet} backbone equipped with FPN \cite{fpn}. The model is illustrated in Figure \ref{fig:architecture}.

\subsection{Object Mask Prior (OMP)}\label{sec:mask-prior}
The OMP functions as an inductive bias capturing a general concept of foreground to improve generalization to weak classes. In the partially supervised learning setup, predicting a general foreground is non-trivial for two main reasons: (1) pixel-wise mask labels are missing for a subset of classes, and (2) in many cases RoIs contain multiple and overlapping instances, making the foreground in a RoI ambiguous. The OMP tackles these issues by highlighting the correct foreground in each RoI, which helps the mask head to learn a more general concept of the foreground, resolve ambiguous RoIs, and generalize it to weak classes.

We create such a prior by extracting the foreground information captured by the box features in the box head. We use the fact that the box classification head learns a representation of the primary class (\ie foreground) for all classes in the dataset. To reveal this foreground, we use class activation maps (CAMs) \cite{cams}, which provide coarse localization maps emphasizing the most discriminative regions the model uses for classification. Consequently, given a correct classification, CAMs are expected to highlight foreground areas corresponding to the primary RoI class. 

To realize CAM calculation, we use a box head with four convolution layers where Global Average Pooling (GAP) is applied on the last convolutional feature map. The resulting vector is processed by linear layers for box classification and regression (see Figure \ref{fig:architecture}). We calculate CAMs with a function $f_{W_{cls}}$ which is a $1\times1$ convolution parameterized with the classification weights $W_{cls}$ as follows:
\begin{align}
    M_{\textit{cam}} = f_{W_{cls}} \left( F_{\textit{box}} \right),
\end{align}
where $F_{box}$ is the last feature map of the box head before GAP. This allows calculating all CAMs efficiently with a single operation while keeping them differentiable. Depending on whether it is training or inference time, we use the ground truth labels or the classes predicted by the box head to select the correct CAM slice from $M_{\textit{cam}}$.

The CAMs of the correct class are added to the corresponding mask features as will be described in the next section. Apart from providing the mask head favorable foreground cues, this also allows the mask gradients backpropagate through the box head. A well known shortcoming of CAMs is that they do not cover the full extent of the objects, but only the minimal area of the most distinctive features. The mask gradients provide the features in the box head mask information, which leads to an increase in the spatial extent of the CAMs allowing them to capture finer details.

As a result, CAMs that receive mask gradients give rise to our OMP. The fact that the OMP originates from the box classification task, which is directly optimized to classify the primary instance in a RoI, provides it with strong foreground cues. This makes our OMP predestined to provide mask head with a general concept of foreground allowing it to resolve ambiguous RoIs and also better generalize to weak classes.

\subsection{Inducing the Prior}\label{sec:mask-head}
After generating the OMP, we aggregate it with the FPN features after the RoIAlign $F_{\textit{fpn}}$ to create object-aware features $F_{\textit{object}}$ as follows:
\begin{align}
    F_{\textit{object}} &= F_{\textit{fpn}} + M_{\textit{cam}} \label{eq:f_object1}\;, 
\end{align}
where $M_{\textit{cam}, k} \in \mathbb{R}^{H, W}$ is added to each channel of its matching RoI $F_{\textit{fpn}, k} \in \mathbb{R}^{D, H, W}$. Before addition, we use bilinear interpolation to adjust $M_{\textit{cam}}$ to the spatial dimensions of $F_{\textit{fpn}}$.

The addition highlights the features in $F_{\textit{fpn}}$ at the spatial locations where the response of the OMP is high. This provides the mask head with explicit foreground information that are embedded by the subsequent convolutional layers in the mask head. This incentivizes the mask head to learn a general concept of foreground for all classes in the dataset.

After the addition, $F_{\textit{object}}$ is processed by a function $f_{\textit{mask}}$ consisting of seven $3\times3$ convolution layers followed by one transposed convolution layer doubling the features spatial resolution and one $1\times1$ convolution performing mask prediction as follows:
\begin{align}
    M_{\textit{mask}} &= f_{\textit{mask}}\left( F_{\textit{object}} \right) \label{eq:f_mask},
\end{align}
where $M_{\textit{mask}}$ is the mask prediction after applying a pixel-wise sigmoid. We use seven convolution layers to achieve a receptive field large enough such that $f_{\textit{mask}}$ operates on the the entire input feature map. Batch normalization \cite{batch_norm} is applied after each $3\times3$ convolution to utilize its stochastic properties to improve generalization. Finally, a pixel-wise binary cross-entropy loss is applied to $M_{\textit{mask}}$ using the available mask labels $M_{\textit{gt}}$ as follows:
\begin{align}
    \mathcal{L}_{Mask} &= \textit{BCE}\left( M_{\textit{mask}}, M_{\textit{gt}} \right).
\end{align}

\section{Experiments}
In Section \ref{sec:exp_setup}, first, the dataset and experimental setup are introduced. Then, in Section \ref{sec:exp_fg_bg}, we provide evidence that instances of weak classes appearing in the background of a RoI during training are learned as background, and a conventional class agnostic mask head has difficulties considering the correct foreground in ambiguous RoIs. Afterwards, Section \ref{sec:gen_nov_class} shows the capabilities of OPMask to generalize to weak classes. Finally, in Section \ref{sec:exp_mask_prior}, we compare our OMP against regular CAMs showing the positive impact of mask gradients updating box features.

\subsection{Experimental Setup}\label{sec:exp_setup}
We conduct our experiments on the COCO dataset \cite{coco}. To realize the partially supervised learning setup, we split the \emph{mask labels} of the $80$ COCO \emph{thing} classes into two subsets. One subset consists of the strongly annotated classes used for training, the other subset consists of the weakly annotated classes used for evaluation and vice versa. Box labels are available for all classes during training. To compare against related work, we mainly focus on dividing the COCO dataset between the $20$ classes of the Pascal voc dataset \cite{pascal_voc} (inside COCO), and the remaining $60$ unique COCO classes. In the following, non-voc $\rightarrow$ voc denotes that the mask head is trained on the non-voc classes and evaluated on voc classes, vice versa the same applies for voc $\rightarrow$ non-voc. For training, we use SGD with Momentum with an initial learning rate of $0.02$ which is linearly warmed up for the first $1000$ iterations \cite{imagenet1hr}. The batch size is set to $16$ and the gradients are stabilized by clipping them at a value of $1.0$. ResNet-50 and ResNet-101 \cite{resnet} with a FPN \cite{fpn} are used as backbones. The implementation is based on PyTorch \cite{pytorch} and Detectron2 \cite{wu2019detectron2}. Further details can be found in the supplementary material.

\textbf{Baseline.} We use a Mask R-CNN with a class agnostic mask head. For a fair comparison, we use the same box head as OPMask and also add batch norm to its mask head. In the following, we call this baseline `Our Mask R-CNN`.

\subsection{Insights on Identifying Foreground in RoIs}\label{sec:exp_fg_bg}
\textbf{Learning classes as background.} 
A class agnostic mask head faces the task of classifying RoI pixels between foreground or background, where pixels that correspond to supervised classes are considered foreground, while all other pixels are regarded as background. The COCO dataset contains complex scenes with cluttered objects, which causes RoIs to often contain more than one instance. Background pixels can either be part of the available supervised classes, belong to weak classes to which we want to generalize, or not be part of any class in the dataset. In the second case, we face the dilemma that the model actively learns to classify features that correspond to weak classes as background. This clearly conflicts with the generalization goal of the partially supervised learning task.

\begin{figure}
\begin{center}
\includegraphics[width=1.0\columnwidth]{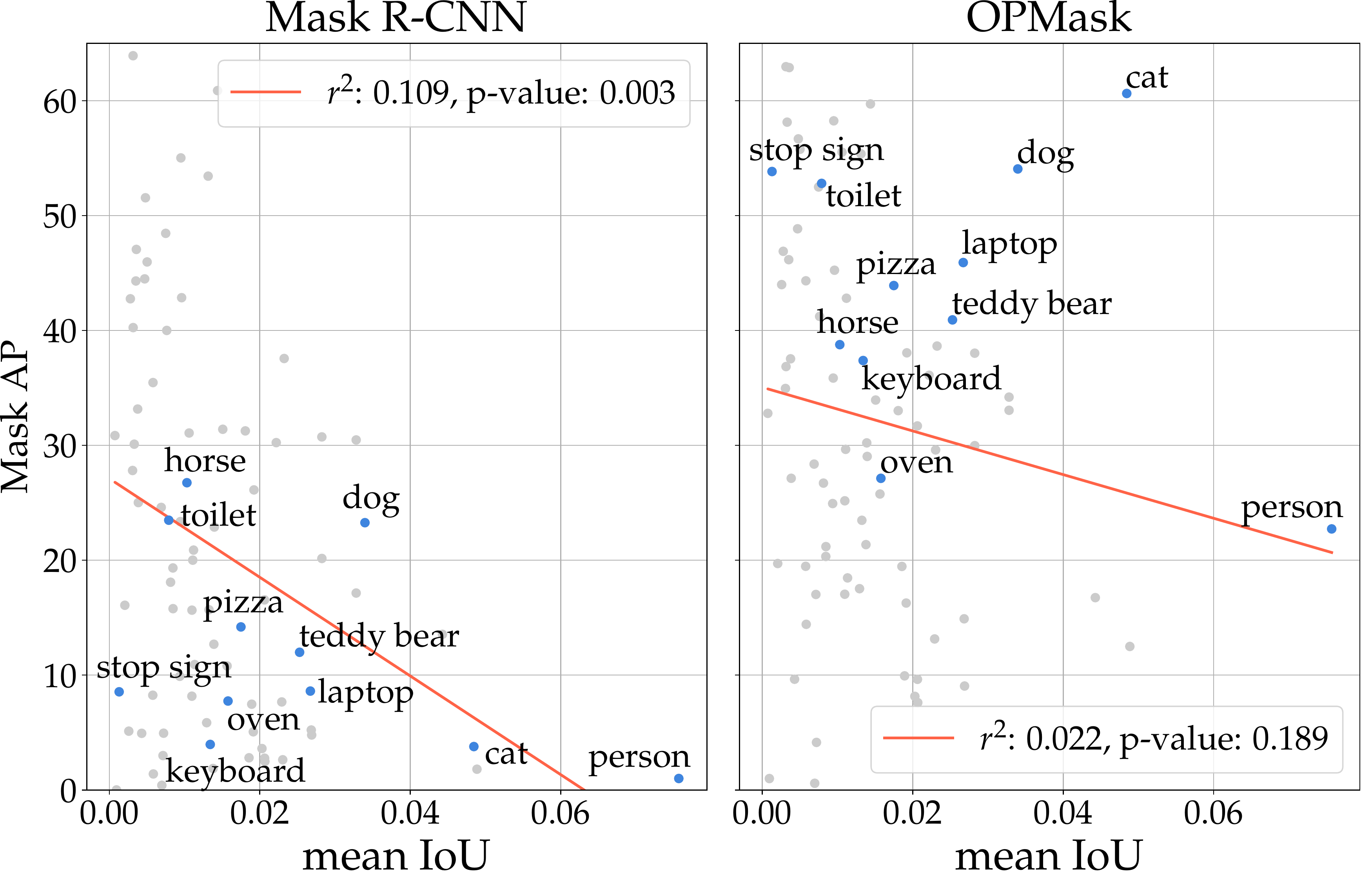}
\end{center}
   \caption{Regressions showing the correlation between box IoU and mask AP of all COCO classes. We compare mask AP scores of our Mask R-CNN baseline (\textit{left}) and OPMask (\textit{right}). The classes with the largest relative improvement are highlighted.}
\label{fig:iou_exp}
\end{figure}

This phenomenon particularly affects classes that frequently interact with other classes and thus appear more often in the background of a mask supervised RoI. To investigate this, we compute the correlation between class overlap and mask AP for weak classes (in voc $\rightarrow$ non-voc and non-voc $\rightarrow$ voc). To approximate the overlap between classes, we compute the IoU of all ground-truth bounding boxes in the COCO dataset. Afterwards, we compute a regression between the mean IoU of each class and its mask AP.

Two regression models are presented in Figure \ref{fig:iou_exp}. The first (\textit{left}) is computed with our Mask R-CNN baseline showing a significant negative correlation between mean IoU and mask AP across all classes ($p=.003 < .01$). This provides evidence for our hypothesis that weak classes appearing in the background of RoIs are actively learned as background during training. The second regression (\textit{right}), computed with OPMask, shows only a weak negative correlation that is not strong enough to reach significance ($p=.189 \nless .01$). At the same time, we see notable improvements for classes with high mean IoU values, which are more likely to appear in the background of other classes RoIs (\eg \textit{person}: $0.99$ to $22.72$ AP, \textit{cat}: $3.77$ to $60.63$ AP). This suggests that the OMP is able to provide the mask head with a general concept of foreground, which counteracts learning these weak classes as background.

\textbf{Resolving ambiguous RoIs.}
Another issue with multiple and possibly overlapping instances is that the primary instance (\ie foreground) of the RoI may be ambiguous. We identify that a conventional mask head has difficulties to locate the foreground in these ambiguous constellations. 

\begin{table}[htb!]
\centering
\resizebox{\columnwidth}{!}{%
\begin{tabular}{l|cc|cc|cc}
                          \multicolumn{1}{c}{}                 & \multicolumn{2}{c}{Full COCO} & \multicolumn{2}{c}{non-voc $\rightarrow$ voc} & \multicolumn{2}{c}{voc $\rightarrow$ non-voc} \\
Model          & Amb.      & $\lnot$Amb.    &  Amb.      & $\lnot$Amb.   & Amb.      & $\lnot$Amb    \\
\Xhline{2\arrayrulewidth} 
Our Mask R-CNN &  15.9     & 36.2     & 10.8    & 27.4   & 6.5    & 19.7   \\
\textbf{OPMask}         &  \textbf{20.5}    & \textbf{38.0}     & \textbf{19.5}    & \textbf{37.2}    & \textbf{17.1}    & \textbf{32.2} \\
\Xhline{2\arrayrulewidth} 
\end{tabular}%
}
\caption{Mask AP of OPMask vs. our Mask R-CNN baseline in ambiguous and non-ambiguous instances. OPMask outperforms our Mask R-CNN baseline in all comparisons, with the largest improvements in ambiguous constellations in the partially supervised setup.}
\label{tab:quantify-ambiguities}
\end{table}

To quantify the effect of ambiguous instances on model performance, we split the COCO validation set into ambiguous and non-ambiguous instances. Since properly quantifying ambiguity is highly non-trivial, we avoid complicated heuristic methods and use the IoU of the box labels as a simple proxy. This allows to capture general trends while putting emphasis on simplicity and reproducibility. We consider instances with a box label IoU $\geq 0.5$ to at least one other instance in an image as ambiguous. Table~\ref{tab:quantify-ambiguities} compares OPMask with our Mask R-CNN baseline having a ResNet-50 backbone in ambiguous and non-ambiguous instances. Both models are trained fully supervised on all COCO classes and partially supervised on the voc vs. non-voc COCO splits. The results show that OPMask performs better than our Mask R-CNN baseline in all comparisons, with the largest improvements in ambiguous constellations. Especially in the partially supervised setting, our Mask R-CNN baseline falls significantly behind OPMask. This suggests that a conventional class agnostic mask head has considerable problems with ambiguous instances, especially when it has to generalize to weak classes. At the same time, OPMask manages to significantly mitigate this drop in the mask AP thanks to the OMP highlighting the correct foreground in the RoIs.

In addition, Figure \ref{fig:fg_bg} provides qualitative comparisons between OPMask and our Mask R-CNN baseline handling ambiguous instances. The results present ambiguous RoIs where our Mask R-CNN baseline falsely predicts background instances as foregrounds. On the other hand, the OMP is able to highlight the foreground instance in the RoI allowing OPMask to correctly segment the instance. All examples are from models trained either in the voc $\rightarrow$ non-voc or non-voc $\rightarrow$ voc setting. Interestingly, the results in the first row are achieved with models trained in the voc $\rightarrow$ non-voc setting, where person is a supervised class and umbrella is a weak class. While the Mask R-CNN incorrectly segments the person, OPMask identifies the umbrella as the primary class and is able to predict an accurate instance mask while generalizing to a weak class. This further emphasizes the quantitative results where our Mask R-CNN baseline performs particularly poorly for ambiguous instances in the partially supervised setting.

\subsection{Generalization to Weakly Annotated Classes}\label{sec:gen_nov_class}
\textbf{Pascal voc vs. non-voc classes.} We present the quantitative results for the voc vs. non-voc splits in Table \ref{tab:main_quant}. The results show that OPMask considerably improves over our Mask R-CNN baseline in all cases. For example, with a ResNet-50 backbone, a significant increase of $10.1$ AP in non-voc $\rightarrow$ voc and $13.0$ AP in voc $\rightarrow$ non-voc is achieved. OPMask also performs better than previous approaches ShapeProp \cite{shapeprop} and ShapeMask \cite{shapemask} in all cases. It is notable that even with a ResNet-50, we achieve better or competitive performance than ShapeMask and ShapeProp that are equipped with the stronger ResNet-101. When comparing to the current state-of-the-art CPMask \cite{cpmask}, we achieve substantial improvements with a ResNet-50 backbone (\eg $3.1$ AP). For the ResNet-50 backbone CPMask only provides results in the voc $\rightarrow$ non-voc setting. Our model with a ResNet-101 backbone achieves competitive performance in non-voc $\rightarrow$ voc (\eg increase of $0.3$ AP), but also slightly worse performance in voc $\rightarrow$ non-voc (\eg $0.8$ AP decrease). It should be noted, however, that unlike us, CPMask uses multi-scale training when equipped with a ResNet-101 backbone, which is known to substantially increase the overall performance of the model. The fact that we significantly outperform CPMask without multi-scale training on a ResNet-50, but only achieve competitive performance on a ResNet-101 demonstrates the magnitude of improvements possible with multi-scale training. Moreover, we emphasize that our model adds only a single matrix multiplication of overhead to the forward pass of a Mask R-CNN. In contrast, CPMask uses an extra boundary-parsing head requiring additional boundary labels and a self-attention based affinity-parsing module that produces expensive $14^2 \times 14^2 \times 256$ feature maps. Further, both modules introduce an additional loss for which gradients must be computed. Thus, in terms of of computing overhead, OPMask emerges as a simpler approach than CPMask.

\begin{table*}[ht!]
\centering
\resizebox{1.0\textwidth}{!}{%
\begin{tabular}{ll|cccccc|cccccc}
                           \multicolumn{2}{c}{} & \multicolumn{6}{c}{non-voc $\rightarrow$ voc: test on voc}   & \multicolumn{6}{c}{voc $\rightarrow$ non-voc: test on non-voc}   \\
Backbone                   & Method                  & $\text{AP}$   & $\text{AP}_{50}$ & $\text{AP}_{75}$ & $\text{AP}_{S}$  & $\text{AP}_{M}$  & $\text{AP}_{L}$  & $\text{AP}$   & $\text{AP}_{50}$ & $\text{AP}_{75}$ & $\text{AP}_{S}$  & $\text{AP}_{M}$  & $\text{AP}_{L}$  \\
\Xhline{3\arrayrulewidth} 
\multirow{5}{*}{R-50-FPN}  & Mask R-CNN \cite{mask_rcnn}      & 23.9 & 42.9 & 23.5 & 11.6 & 24.3 & 33.7 & 19.2 & 36.4 & 18.4 & 11.5 & 23.3 & 24.4 \\
                           & Our Mask R-CNN  & 26.4 & 46.4 & 26.7 & 14.2 & 26.4 & 36.5 & 18.9 & 35.5 & 18.4 & 12.4 & 22.8 & 22.9 \\
                           & Mask\textsuperscript{\textit{X}} R-CNN \cite{ltse} & 28.9 & 52.2 & 28.6 & 12.1 & 29.0 & 40.6 & 23.7 & 43.1 & 23.5 & 12.4 & 27.6 & 32.9 \\
                           & Mask R-CNN w/ ShapeProp \cite{shapeprop}   & 34.4 & 59.6 & 35.2 & 13.5 & 32.9 & 48.6 & 30.4 & 51.2 & 31.8 & 14.3 & 34.2 & 44.7 \\
                           &  CPMask \cite{cpmask} & - & - & - & - & - & - & 28.8 & 46.1 & 30.6 & 12.4 & 33.1 & 43.4 \\
                           & \textbf{OPMask} & \textbf{36.5} & \textbf{62.5} & \textbf{37.4} & \textbf{17.3} & \textbf{34.8} & \textbf{49.8} &  \textbf{31.9} & \textbf{52.2} & \textbf{33.7} & \textbf{16.3} & \textbf{35.2} & \textbf{46.5}  \\
\Xhline{2\arrayrulewidth} 
\multirow{7}{*}{R-101-FPN} & Mask R-CNN \cite{mask_rcnn} & 24.7 & 43.5 & 24.9 & 11.4 & 25.7 & 35.1 & 18.5 & 34.8 & 18.1 & 11.3 & 23.4 & 21.7 \\
                           & Our Mask R-CNN & 27.7 & 48.0 & 28.2 & 13.6 & 28.6 & 38.0 & 21.0 & 39.2 & 20.5 & 13.5 & 26.4 & 23.9 \\  
                           & Mask\textsuperscript{\textit{X}} R-CNN \cite{ltse} & 29.5 & 52.4 & 29.7 & 13.4 & 30.2 & 41.0 & 23.8 & 42.9 & 23.5 & 12.7 & 28.1 & 33.5 \\
                           & ShapeMask \cite{shapemask}              & 33.3 & 56.9 & 34.3 & 17.1 & 38.1 & 45.4 & 30.2 & 49.3 & 31.5 & 16.1 & 38.2 & 28.4 \\
                           & Mask R-CNN w/ ShapeProp \cite{shapeprop} & 35.5 & 60.5 & 36.7 & 15.6 & 33.8 & 50.3 & 31.9 & 52.1 & 33.7 & 14.2 & 35.9 & 46.5 \\
                           & CPMask \cite{cpmask}  & 36.8 & 60.5 & \textbf{38.6} & \textbf{17.6} & \textbf{37.1} & \textbf{51.5} & \textbf{34.0} & \textbf{53.7} & \textbf{36.5} & \textbf{18.5} & \textbf{38.9} & \textbf{47.4} \\
                           & \textbf{OPMask} & \textbf{37.1} & \textbf{62.5} & 38.4 & 16.9 & 36.0 & 50.5 & 33.2 & 53.5 & 35.2 & 17.2 & 37.1 & 46.9 \\
\Xhline{2\arrayrulewidth}
\end{tabular}%
}
\caption{Comparing OPMask with the state-of-the-art in the partially supervised instance segmentation setup on COCO. OPMask outperforms our Mask R-CNN baseline as well as our related work Mask\textsuperscript{\textit{X}} R-CNN, ShapeMask and ShapeProp with both backbones. Further, OPMask performs better than CPMask on a ResNet-50 and remains competitive on a ResNet-101. non-voc $\rightarrow$ voc denotes that the mask head is trained on the non-voc classes and evaluated on voc classes, vice versa the same applies for voc $\rightarrow$ non-voc. 
}
\label{tab:main_quant}
\end{table*}

\begin{figure*}
\begin{center}
\includegraphics[width=0.8\linewidth]{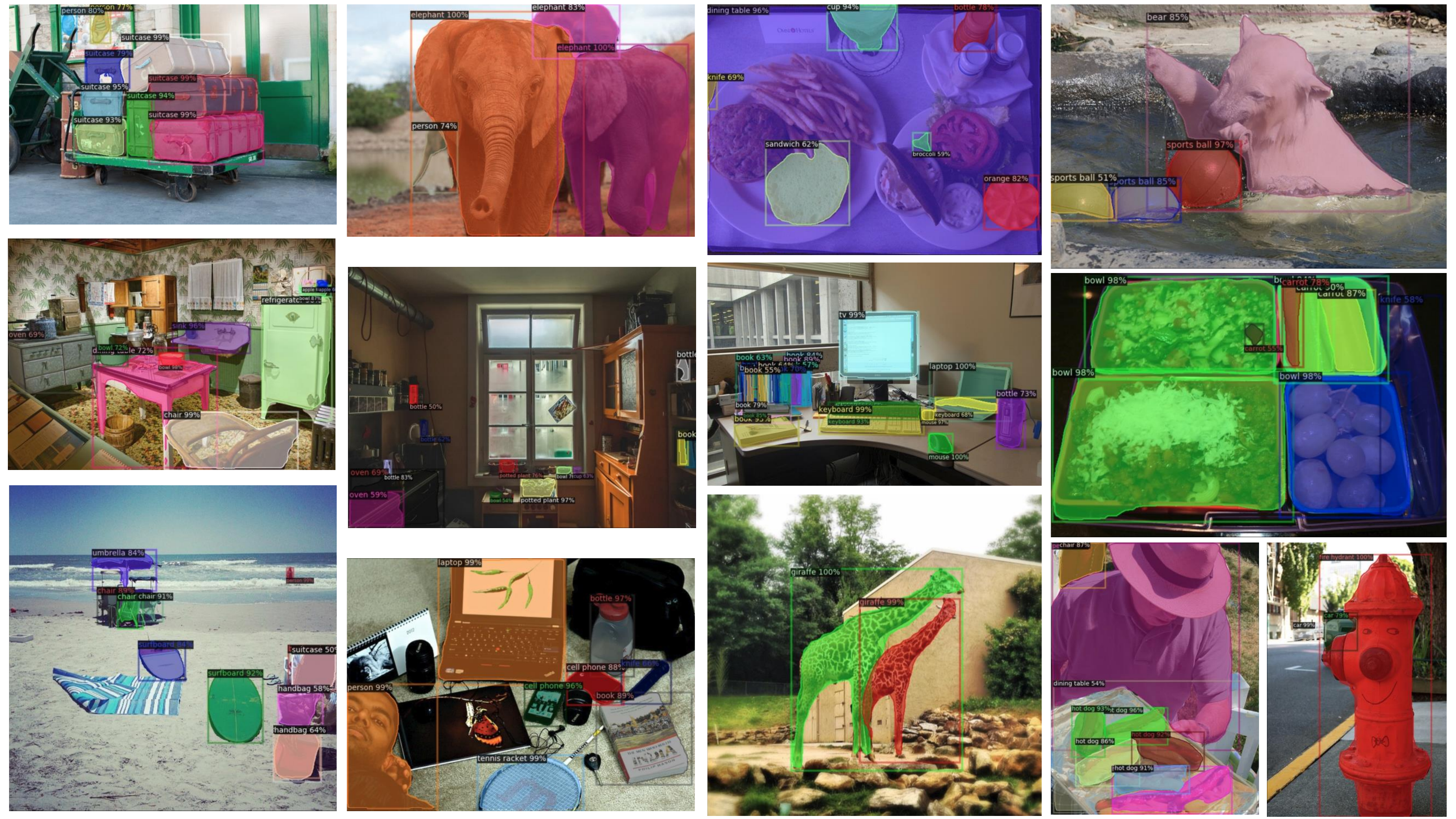}
\end{center}
   \caption{Qualitative results on COCO using the voc $\rightarrow$ non-voc split for training. This shows the ability of OPMask to predict precise instance masks for objects of weak classes across different scenes, and various object sizes and appearances.}
\label{fig:main_qual}
\end{figure*}

\textbf{Qualitative Results.}
\begin{figure}
\begin{center}
\includegraphics[width=0.9\columnwidth]{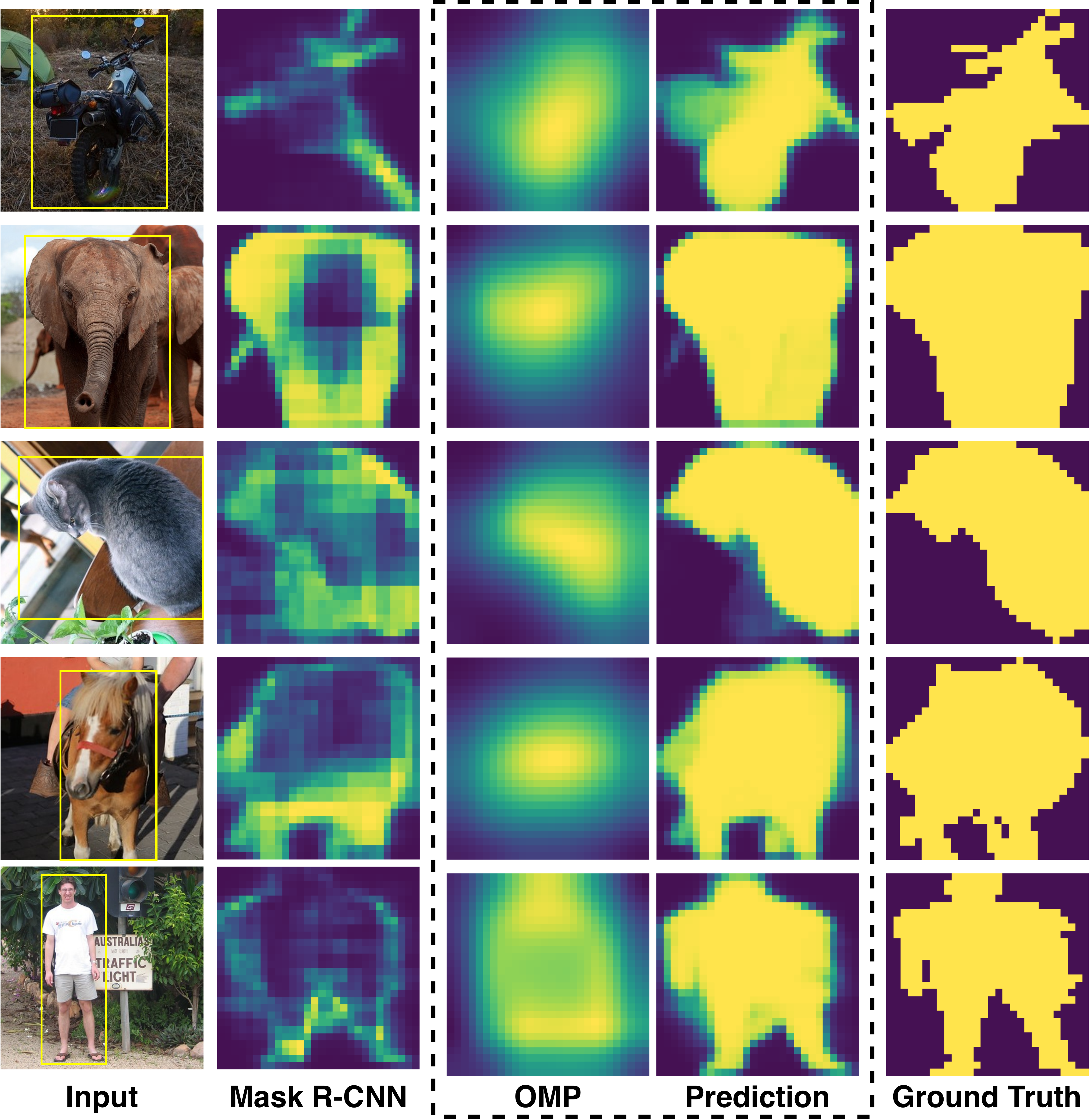}
\end{center}
   \caption{The Mask R-CNN baseline produces perforated, incomplete or missing masks. OPMask driven by the OMP is able to accurately segment each instance of a weak class.}
\label{fig:prior_and_preds}
\end{figure}
In Figures \ref{fig:fg_bg} and \ref{fig:prior_and_preds}, we provide qualitative insights into how the OMP steers mask predictions and improves generalization to weak classes. Each example shows a weak class in either the voc $\rightarrow$ non-voc or non-voc $\rightarrow$ voc setup. Next to the OMP and mask prediction of OPMask, our Mask R-CNN baseline predictions are presented. The results show that the OMPs properly identify and highlight the primary instances in the RoIs while covering most of the objects' spatial extend. Furthermore, we realize that our coarse prior is sufficient to enable the mask head to generalize to a refined mask. This underlines our hypothesis that it is of particular importance to provide the class agnostic mask head with a general concept of foreground across all classes.
Finally, Figure \ref{fig:main_qual} presents a number of COCO images with overplayed mask predictions produced in the voc $\rightarrow$ non-voc setup. The results show OPMask's ability to generate precise predictions for instances of weak classes across different scenarios and object sizes. All examples in this section are achieved with models equipped with a ResNet-101.

\begin{figure}
\begin{center}
\includegraphics[width=1.0\columnwidth]{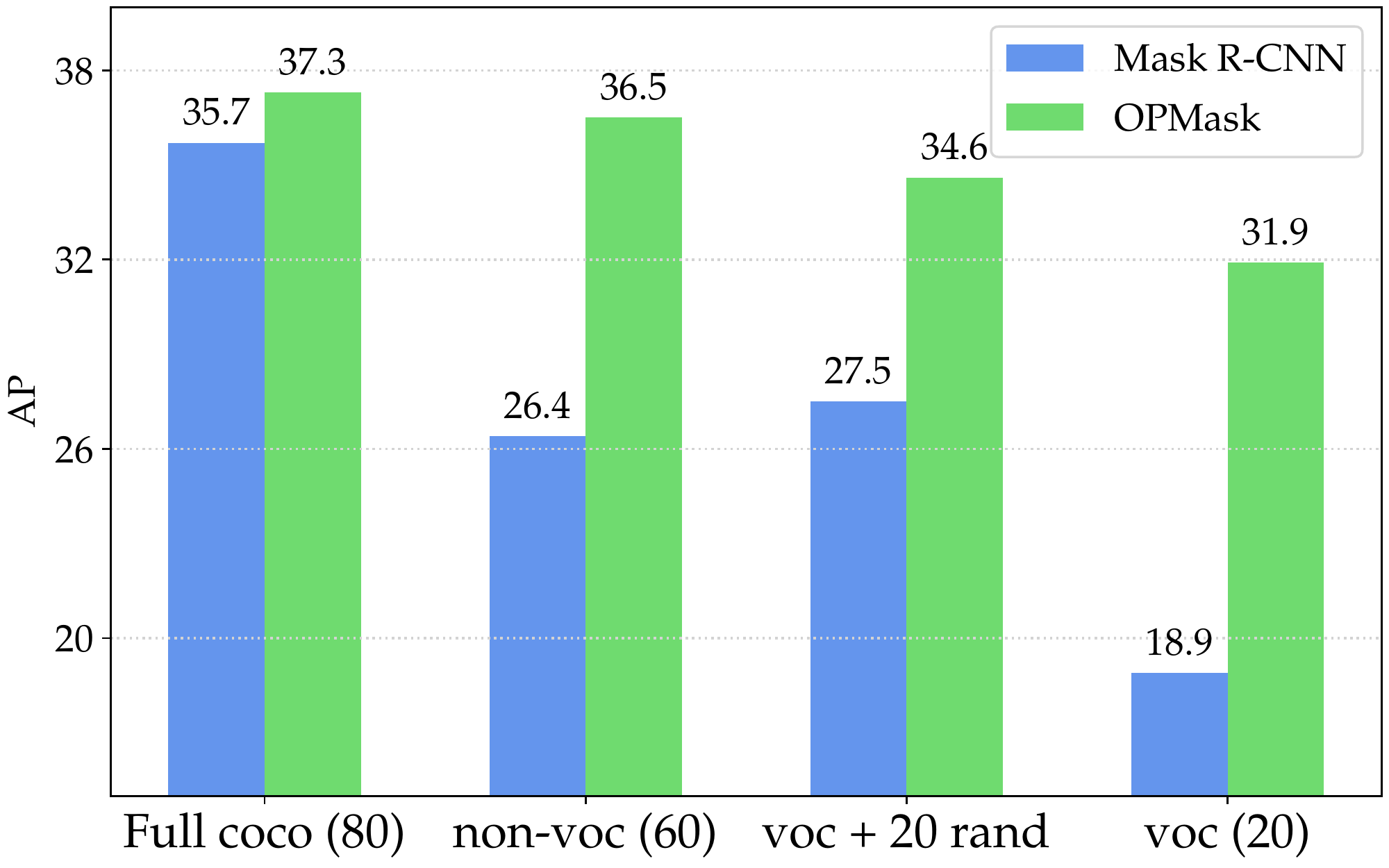}
\end{center}
   \caption{Performance of OPMask on varying numbers of supervised classes. It significantly improves over our Mask R-CNN baseline across all class splits, including the fully supervised setup.}
\label{fig:less_classes}
\end{figure}

\textbf{Strongly vs weakly supervised class ratios.}
To provide a better overview of OPMask's generalization abilities, we evaluate its performance on different splits of strongly and weakly annotated classes. To create the $40$ class split, we start with the $20$ Pascal voc \cite{pascal_voc} classes and randomly add another $20$ classes from the non-voc split. Figure \ref{fig:less_classes} shows that OPMask consistently improves over our Mask R-CNN baseline while demonstrating stable performance across all class splits. Even in the fully supervised setup (Full coco), where a Mask R-CNN remains competitive with other state-of-the-art instance segmentation methods, OPMask achieves better performance ($+1.6$ AP). We attribute these improvements to the fact that even with full supervision, the OMP helps the class agnostic mask head resolve ambiguous RoIs, which in return improves OPMask predictions. This shows that our OMP remains beneficial when strong annotations are available for all classes.

\subsection{Refining the Object Mask Prior}\label{sec:exp_mask_prior}
A simple CAM as the OMP might do a reasonable job, though a better prior is expected to lead to a better segmentation result. To improve our OMP, we let mask gradients backpropagate through the box head, which augments the box features with mask information. This enhances the CAMs by increasing their spatial extent and allowing the OMP to cover larger parts of the objects. The resulting refinement further improves the final mask AP by $1.1$ points in non-voc $\rightarrow$ voc with ResNet-50 backbone. To investigate the improvement of the prior, we compare the mask AP of the OMP with vanilla CAMs on the COCO validation set. We compare against a Faster R-CNN and a Mask R-CNN with the same box head as OPMask. In Table \ref{tab:cam_refine}, $\text{AP}$ and $\text{AP}_{50}$ results of voc vs. non-voc class splits are provided. Since the Faster R-CNN does not receive any mask gradients, it is only trained and evaluated on all classes.

\begin{table}[ht!]
\centering
\resizebox{1.0\columnwidth}{!}{%
\begin{tabular}{l|c|cc|cc|cc}
                          \multicolumn{2}{c}{}  & \multicolumn{2}{c}{test on all} & \multicolumn{2}{c}{test on voc}  & \multicolumn{2}{c}{test on non-voc}   \\
Model           &   train set      & $\text{AP}$  & $\text{AP}_{50}$ & $\text{AP}$  & $\text{AP}_{50}$ & $\text{AP}$  & $\text{AP}_{50}$  \\
\Xhline{3\arrayrulewidth} 
Faster R-CNN    &  all      & 0.2 & 1.0 &  0.3 & 2.5 &  0.1 & 0.6 \\
Mask R-CNN      &  all      & 0.2 & 1.3 & 0.3 & 2.3 & 0.1 & 1.0  \\
OPMask        &  all      & \textbf{8.8} & \textbf{34.1} & \textbf{9.9} & \textbf{40.4} & \textbf{8.4} & \textbf{32.0}  \\
\Xhline{2\arrayrulewidth}
Mask R-CNN      &  voc      & 0.3 & 1.9 & 0.6 & 4.0 & 0.2 & 1.0 \\
OPMask        &  voc      & \textbf{5.0} & \textbf{21.8} &  \textbf{9.9} & \textbf{38.5} &  \textbf{3.3} & \textbf{16.3} \\
\Xhline{2\arrayrulewidth}
Mask R-CNN      &  non-voc  & 0.2 & 1.5 & 0.4 & 2.8  & 0.2 & 1.0 \\
OPMask        &  non-voc  & \textbf{8.0} & \textbf{31.5} & \textbf{7.0} & \textbf{32.2} & \textbf{8.3} & \textbf{31.2}  \\
\Xhline{2\arrayrulewidth} 
\end{tabular}%
}
\caption{Quantitatively comparing our OMP with CAMs of a Faster R-CNN and Mask R-CNN. The results show that our OMP is able to cover larger parts of the objects than conventional CAMs.}
\label{tab:cam_refine}
\end{table}

\begin{figure}
\begin{center}
\includegraphics[width=0.9\columnwidth]{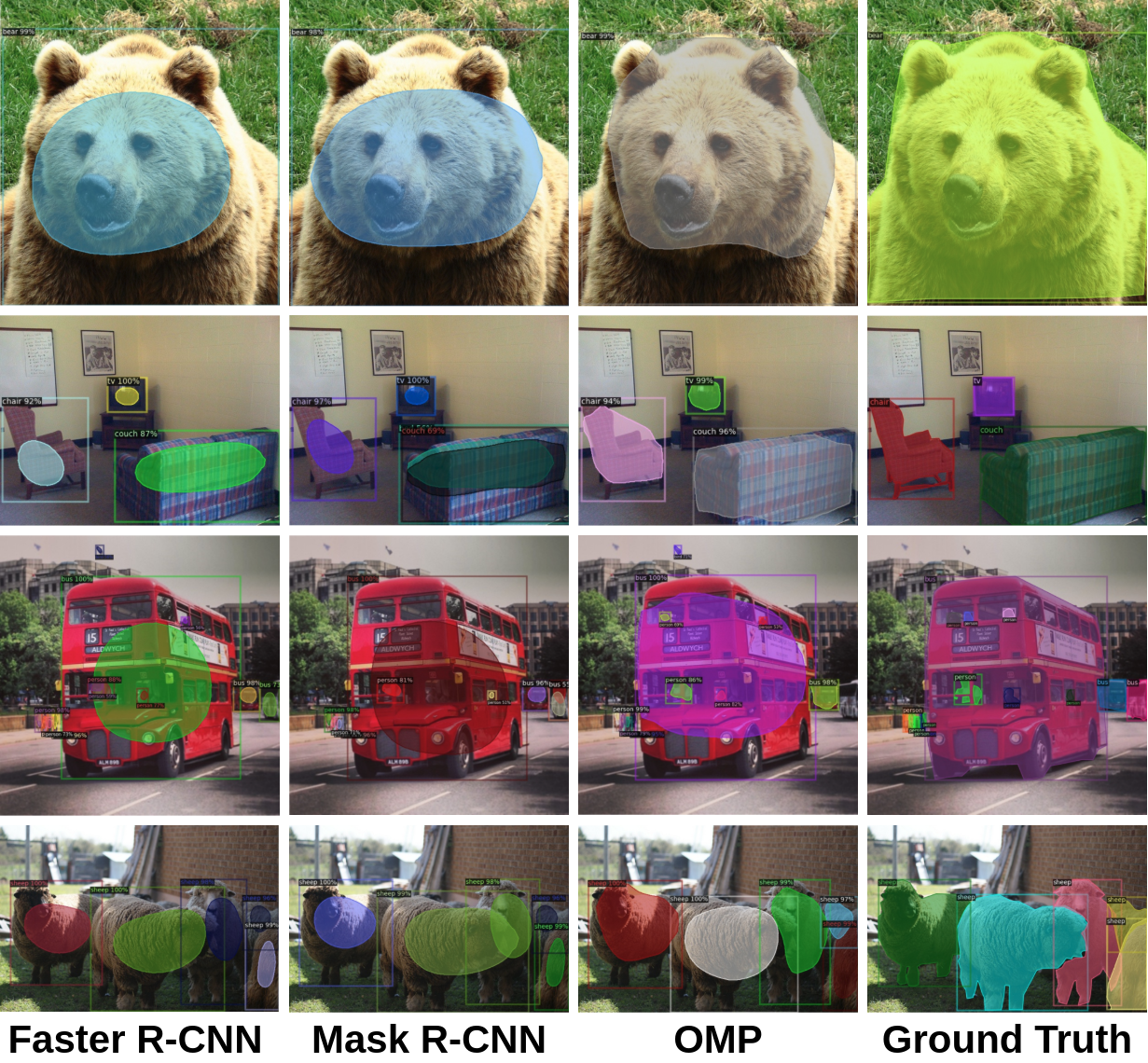}
\end{center}
  \caption{Comparing our OMP with CAMs from a Faster R-CNN and Mask R-CNN on COCO images. We see that our OMP is able to cover larger parts of the objects than regular CAMs.}
\label{fig:cam_quality}
\end{figure}

The results show that the OMP is significantly better than the CAMs of Faster R-CNN and Mask R-CNN. This underlines the positive influence of mask gradients on box head features and consequently on the OMP. The low AP values of the CAMs generated by Faster R-CNN and Mask R-CNN are caused by the fact that they often do not surpass the pixel-wise IoU threshold (i.e. $\geq 0.5$), and are mostly considered negatives. The Mask R-CNN, where the backbone features are augmented with mask gradients, does not show significant improvements over the Faster R-CNN. This suggests that for CAM refinement, mask gradients should impact the features that are directly used to calculate the CAM activations. Finally, Figure \ref{fig:cam_quality} demonstrates qualitative improvements of CAMs on a number of COCO images.

\section{Conclusion}
We proposed OPMask, a novel approach to partially supervised instance segmentation. OPMask introduces an object mask prior (OMP) that helps its class agnostic mask head to learn a general concept of foreground, resolve ambiguous RoIs and generalize to weak classes. Our research pointed out two major problems hindering a class agnostic mask head to generalize to weak classes. First, instances of weak classes appearing in the background of a mask supervised RoI during training are learned as background by the model. Second, in ambiguous RoIs that contain multiple and possibly overlapping instances, the mask head has difficulties to consider the correct foreground. We demonstrated that both problems can be vastly alleviated with our OMP that highlights foreground across all classes by leveraging the information from the box head. Finally, we showed that OPMask significantly improves over our Mask R-CNN baseline and achieves competitive performance with the state-of-the-art, while offering a much simpler architecture with less computational overhead.

{\small
\bibliographystyle{ieee_fullname}
\bibliography{egbib}
}

\end{document}